\documentclass[lettersize,journal]{IEEEtran}
\usepackage{amsmath,amsfonts}
\usepackage{algorithmic}
\usepackage{algorithm}
\usepackage{array}
\usepackage[caption=false,font=normalsize,labelfont=sf,textfont=sf]{subfig}
\usepackage{textcomp}
\usepackage{stfloats}
\usepackage{url}
\usepackage{verbatim}
\usepackage{graphicx}
\usepackage{cite}
\begin{document}

\title{Facial Expression Generation Aligned with Human Preference \\for Natural Dyadic Interaction}

\author{Xu Chen, Rui Gao, Xinjie Zhang, Haoyu Zhang, Che Sun, Zhi Gao, Yuwei Wu, Yunde Jia}



\maketitle

\begin{abstract}
Achieving natural dyadic interaction requires generating facial expressions that are emotionally appropriate and socially aligned with human preference. Human feedback offers a compelling mechanism to guide such alignment, yet how to effectively incorporate this feedback into facial expression generation remains underexplored. In this paper, we propose a facial expression generation method aligned with human preference by leveraging human feedback to produce contextually and emotionally appropriate expressions for natural dyadic interaction. A key to our method is framing the generation of identity-independent facial expressions as an action learning process, allowing human feedback to assess their validity free from visual or identity bias. We establish a closed feedback loop in which listener expressions dynamically respond to evolving conversational cues of the speaker. Concretely, we train a vision-language-action model via supervised fine-tuning to map the speaker’s multimodal signals into controllable low-dimensional expression representations of a 3D morphable model. We further introduce a human-feedback reinforcement learning strategy that integrates the imitation of high-quality expression response with critic-guided optimization. Experiments on two  benchmarks demonstrate that our method effectively aligns facial expressions with human preference and achieves superior performance.
\end{abstract}

\IEEEpubidadjcol

\section{Introduction}

\begin{figure}[!b]
    \centering
    \includegraphics[width=\linewidth]{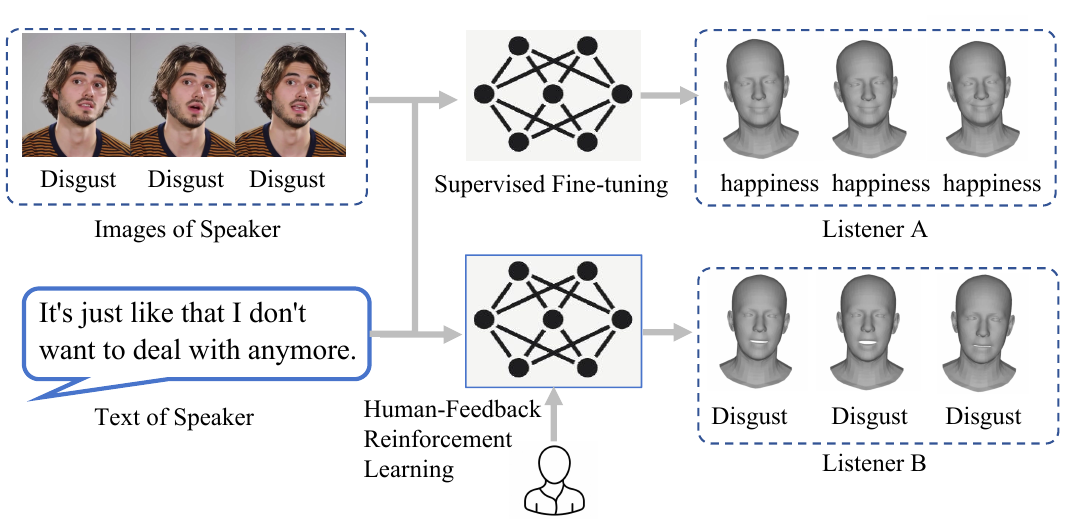}
    \caption{Examples of facial expression generation with (Listener B) and without (Listener A) human preference alignment. Listener B, aligned via human feedback, generates a disgust expression consistent with the Speaker’s emotion, whereas the unaligned Listener A generates an inconsistent happiness expression. }
    \label{fig:first}
\end{figure}
Facial expression generation in dyadic interaction aims to generate the listener’s facial responses based on the speaker’s multimodal cues, including speech, language, and visual dynamics, to enable more human-like expression interactions, which has drawn increasing attention in human–computer interaction~\cite{luo2024reactface,ZhuZRHLG25}. Recent advances~\cite{curto2021dyadformer,ng2022learning,ZhuZRHLG25} in dyadic facial expression generation leverage deep generative models, such as diffusion networks~\cite{croitoru2023diffusion,ho2020denoising} and generative adversarial networks~\cite{goodfellow2014generative}, to generate listeners' facial expressions conditioned on the speaker’s multimodal cues, and have achieved impressive performance.  However, few studies explicitly consider expression generation aligned with human preference, which requires that these expressions be consistent with human social norms and emotional expectations. This alignment is crucial, as evidence from both social psychology~\cite{tisserand2024rejecting} and conversation analysis~\cite{herbert2021deal} suggests that expressions misaligned with contextual or social norms (e.g., inappropriate emotion or disregard for ``preference organization'' in turns) can disrupt interactional flow and reduce user acceptance. A typical example of this limitation can be seen in Listener A in Figure~\ref{fig:first}, where the listener generates a happiness expression while the speaker conveys disgust, likely leading to conversational dissonance and a breakdown in social rapport.  In this paper, we focus on facial expression generation aligned with human preference, aiming to ensure that listener expressions are emotionally appropriate and socially compatible with human expectations.

A natural way to achieve alignment with human preference is to incorporate human feedback into the learning process, allowing the model to align expression generation with human judgments. However, directly using the feedback to generate facial expression in dyadic interaction is non-trivial. First,  generated expressions are often entangled with identity and appearance, making it challenging to obtain unbiased human feedback that truly reflects the expression quality. This entanglement causes human raters to confuse visual realism or appeal with the quality of expressions. Consequently, the human feedback would become sensitive to identity or appearance rather than to the objective expressions, limiting its reliability as a learning signal for human-preference alignment.  Second, achieving continuous alignment is also challenging, as it requires models to dynamically adjust expressions in response to evolving speaker behaviors and conversational dynamics, whereas most generative methods operate in an open-loop manner that produces unchanged outputs without adaptive feedback. 
\IEEEpubidadjcol

In this paper, we propose a novel method of facial expression generation aligned with human preference by leveraging human feedback to produce contextually and emotionally appropriate listener expressions for natural dyadic interaction. The core idea is to frame expression generation as an action learning process in an identity-independent space, enabling human feedback to reliably assess expression quality without visual or identity bias. We also establish a closed feedback loop in which listener expressions dynamically adapt to the speaker’s evolving multimodal cues, thereby enabling continuous alignment.  Specifically, we use a vision–language–action model and use a supervised fine-tuning strategy to map the speaker’s facial motion and language cues into controllable low-dimensional expression actions represented on a 3D morphable model. We further introduce a human-feedback reinforcement learning strategy to integrate imitation of desirable expressions with critic-guided refinement, enabling iterative optimization of alignment quality. By introducing human-feedback reinforcement learning, our listener (exemplified by Listener B in  in Fig~\ref{fig:first}) generates aligned expressions, such as generating disgust expressions in response to the speaker's disgust emotion, which significantly improves interactional coherence and perceived social appropriateness.

As demostrated in the bottom row of Figure~\ref{fig:first}, our method achieves superior alignment with the speaker's "disgust" cues, fostering a more natural and socially resonant dyadic interaction. 

We evaluate our method by generating listener expressions on two widely used datasets: L2L-trevor~\cite{NgSKKDG23}, and Realtalk~\cite{geng2023affectivefacesgoaldrivendyadic}. Experimental results show the superiority of our method.

The main contributions of this work are summarized as follows:
\begin{itemize}
\item 

To the best of our knowledge, we are the first to explicitly use human feedback in a closed-loop manner to align facial expression generation with human preference for natural dyadic interaction.  This alignment ensures that listener responses are not only visually natural but contextually and emotionally appropriate.

\item We propose a facial expression generation method aligned with human preference that effectively integrates human feedback by formulating expression generation as an action learning process in an identity-independent space, enabling the model to learn expressions that reflect human preferences without bias.

\end{itemize}

\section{Related Work}

\subsection{3D Talking Head Generation}
Existing 3D talking head generation methods have primarily focused on single/dyadic-role and audio-driven facial animation synthesis. Early works focus on lip-synchronized facial animations aligning to the speaker's audio~\cite{CaoTFP05,HavellRSAMH12}. Recent works such as FaceFormer~\cite{FanLSWK22} and CodeTalker~\cite{XingXZC0W23} introduce the Transformer architectures to capture long-term dependency of multi-model information for generating facial meshes with geometric structure, achieving high-precision lip synchronization and realistic 3D talking head generation. To further improve generation quality and efficiency, recent research has made improvements from both architectural and training strategy perspectives, such as SadTalker~\cite{ZhangCWZSGSW23}, TalkingMachines~\cite{LOWW2025}, and so on. Besides, MultiTalk~\cite{SungBinCSHJNO24} constructs a cross-lingual dataset and improves the accuracy of mouth shape synchronization under different language inputs by introducing multilingual style embeddings.   ConsistentAvatar~\cite{YangZ0Q024} introduces a diffusion-based neural render to generate temporal, 3D-aware, and expression-consistent talking head avatars.  DualTalk~\cite{PengFWW0HF25} extends audio-driven 3D talking head generation to dyadic interaction modeling, and introduces a dual-speaker joint encoder and interaction module to simulate multi-round conversations involving both speaking and listening roles.

Prior to our work, Avatar Forcing~\cite{ki2026forcing} introduces preference alignment into 3D talking head generation, enhancing expressiveness through direct preference optimization (DPO). However, its alignment is constrained by reliance on synthetic proxy data rather than real human feedback, which limits its ability to capture the subtle and complex social norms as well as  emotional expectations in human-to-human interaction. In contrast, our work establishes a closed-loop human-feedback framework that directly generates and refines facial expressions in an identity-independent action space, enabling human preference alignments from real human feedback.

\subsection{Listener Modeling}
The goal of listener modeling is to generate listeners' non-verbal feedback (such as nodding, smiling) based on the speaker's multimodal inputs (speech, language, vision), or to achieve dynamic switching between both roles. In the listener feedback generation task, L2L~\cite{NgSKKDG23} first uses VQ-VAE to discretely encode head motion sequences, generating non-deterministic listening reactions. Subsequently, RLHG~\cite{ZhouBZYZM22} proposes the ViCo dataset and establishes a baseline model based on sequential decoding. Further, ELP~\cite{SongYJDX23} and CustomListener~\cite{LiuGZLAY24} enhance reaction diversity and controllability from the perspectives of emotional priors and text control, respectively. In the domain of dyadic interaction modeling, some prominent methods have emerged, e.g., DIM~\cite{TranCSS24} and INFP~\cite{ZhuZRHLG25}, etc.  DIM~\cite{TranCSS24} generates non-deterministic facial motions in dyadic interactions by utilizing a VQ-VAE to encode actor motions into discrete latent representations. INFP~\cite{ZhuZRHLG25}  introduces an audio-driven interactive head generation framework that operates through a two-stage pipeline.

Existing listener modeling methods (e.g., L2L~\cite{NgSKKDG23}, DIM~\cite{TranCSS24}, ReactFace~\cite{LuoSXSGSG25}) have achieved remarkable progress. Most of them rely on imitation learning from real-world datasets and treat all samples in a dataset as equally ``correct'', failing to distinguish between high-quality social interactions and neutral or inattentive listener states. In contrast, our method explicitly learns from human-preference feedback to generate listener expressions that are not only realistic but also socially and emotionally appropriate, thereby moving beyond mere imitation towards optimized human alignment.
\begin{figure*}[t]
    \centering
    \includegraphics[width=\linewidth]{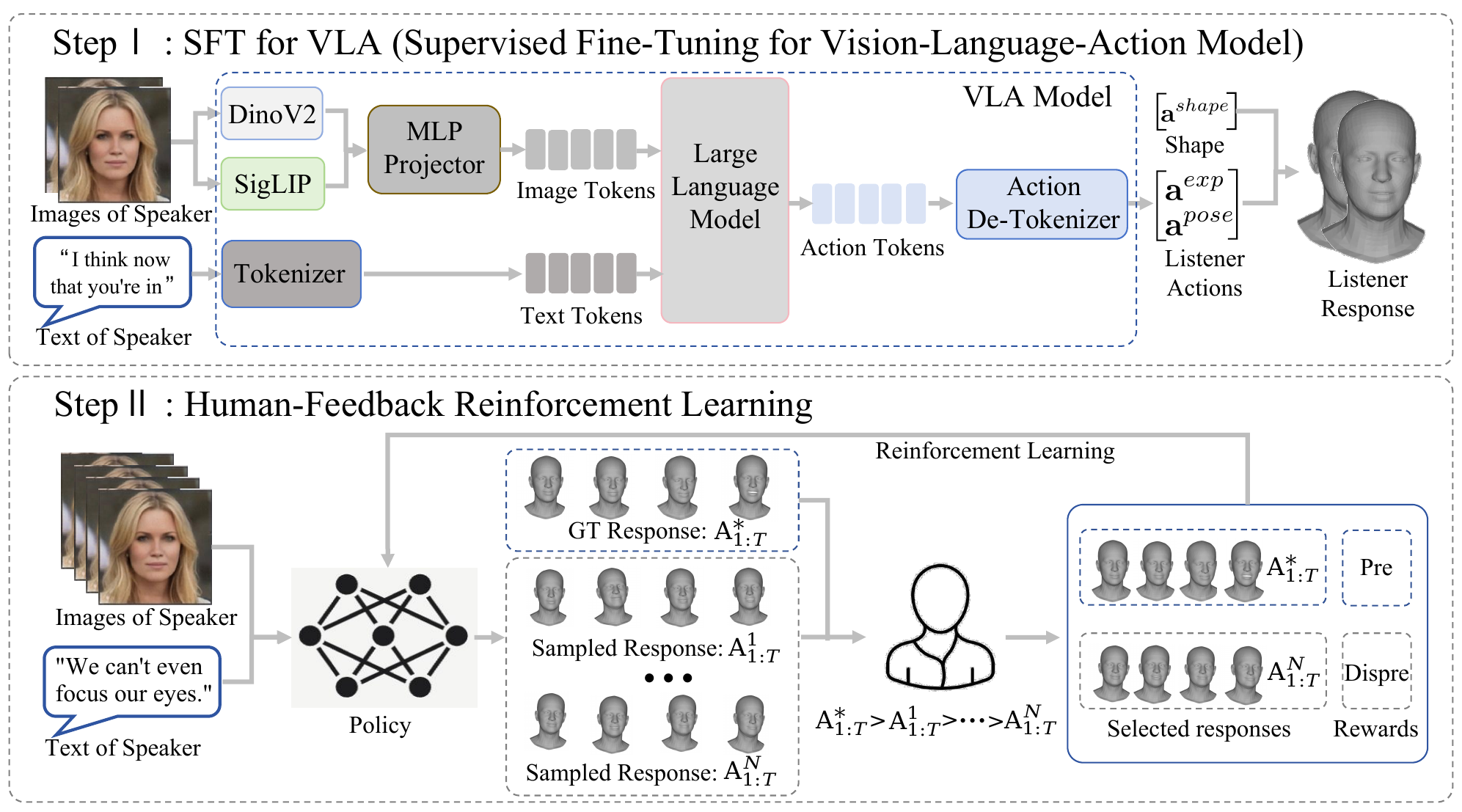}
    \caption{Overall framework of our method.}
    \label{fig:framework}
\end{figure*}

\subsection{Preference Alignment in Generative Tasks}
Preference alignment has emerged as a key problem for steering generative models toward outputs that better reflect human judgment, and has achieved foundational success in natural language processing~\cite{00010LYHLW24,Shani0CLCZNKPSH24,ChaudhariAMRKNDS26}. The study of preference alignment has subsequently been extended to other modalities. In audio and speech synthesis, methods such as BATON~\cite{Liao2024baton} and SpeechAlign~\cite{ZhangLLZWZQ24} have demonstrated that incorporating human-centric utility metrics can enhance prosodic quality and emotional expressiveness beyond the limits of conventional maximum-likelihood training. A similar trend is observed in the vision-language domain, where models like SHAPE~\cite{chen2025SHAPE} employ preference-based tuning to improve the relevance and coherence of generated descriptions. These advances underscore the effectiveness of preference-driven learning across modalities, while the alignment of human preferences in facial expression generation remains unexplored. A core challenge is to obtain unbiased human feedback that truly reflects the expression quality. Our method addresses this challenge by framing expression generation as an action learning process in an identity-independent space, thereby optimizing facial expressions with human preference independent of identity.

\section{Method}

\subsection{Problem Formulation}

We formulate the dyadic facial expression generation task as a sequential decision-making problem in a closed-loop interaction strategy. Given a sequence of the speaker's multimodal inputs, denoted as $\mathbf{S}_{1:t} = \{\mathbf{s}_1, \mathbf{s}_2, \dots, \mathbf{s}_t\}$, where each state $\mathbf{s}_i = (\mathbf{I}_i, \mathbf{L}_i)$ comprises visual frames $\mathbf{I}_i$ and language content $\mathbf{L}_i$, our goal is to learn a policy $\pi_\theta$ that generates appropriate listener facial parameters $\mathbf{A}_t \sim \pi_\theta(\cdot|\mathbf{S}_{1:t})$ at each time step.

The facial parameters $\mathbf{A}_t = [\mathbf{a}_t^{\text{exp}}; \mathbf{a}_t^{\text{pose}}]$ consist of expression coefficients $\mathbf{a}_t^{\text{exp}} \in \mathbb{R}^{D^{\text{exp}}}$ and head pose parameters $\mathbf{a}_t^{\text{pose}} \in \mathbb{R}^{D^{\text{pose}}}$, and they are used to render the 3D face mesh $\mathcal{M}_t$ through the FLAME model~\cite{LiBBL017} while maintaining the fixed identity parameters $\mathbf{a}^{\text{shape}}$, given by
\begin{equation}
    \mathcal{M}_t = \text{FLAME}(\mathbf{a}^{\text{shape}}, \mathbf{a}_t^{\text{exp}}, \mathbf{a}_t^{\text{pose}}). \label{equ:flame}
\end{equation}

This formulation directly serves our goal of human-preference alignment. By defining the listener’s response as an  policy output in a sequential interaction, our method is suitable for feedback-driven learning, allowing the model to adapt based on human judgments. Moreover, the use of identity-agnostic facial parameters as actions focuses the policy on learning transferable expression dynamics and establishes a  disentangled space for collecting human feedback that targets social appropriateness rather than visual appearance.

\subsection{Overview of the Method}
We propose a facial expression generation method aligned with human preferences. 
As illustrated in Figure~\ref{fig:framework}, our method operates in two stages.
In the first stage, the vision-language-action (VLA) model takes the speaker's images $\{\mathbf{I}_1, \mathbf{I}_2, \dots, \mathbf{I}_T\}$ and the corresponding text content $\{\mathbf{L}_1, \mathbf{L}_2, \dots, \mathbf{L}_T\}$ as inputs to predict the listener's facial expression actions $\mathbf{A}_t=[\mathbf{a}_t^{\text{exp}}; \mathbf{a}_t^{\text{pose}}]$. During this stage, the model is trained via Supervised Fine-Tuning (SFT) to imitate the ground-truth listener actions, thereby acquiring a foundational capability for facial expression synthesis.
In the second stage, we introduce a Human-Feedback Reinforcement Learning strategy. We utilize the policy (i.e., the VLA model trained in the SFT stage) to sample $N$ candidate listener actions $\{\textbf{A}_{1:T}^1,...,\textbf{A}_{1:T}^N\}$. Subsequently, these sampled actions are rendered into visual listener responses to facilitate human assessment. These generated responses, together with the ground-truth (GT) response $\textbf{A}_{1:T}^*$, are then evaluated and ranked by human annotators. According to these rankings, we designate the highest-scored response as the human-preferred (Pre) sample and the lowest-scored response as the human-dispreferred (Dispre) sample. Finally, we employ the Direct Preference Optimization (DPO) algorithm to optimize the policy using these selected preference pairs.

\subsection{Vision-Language-Action Model}

Our vision-language-action model consists of three key components including multimodal input encoding, action de-tokenizer, and  a large language model (LLM) backbone.
In this work, we use the 7B-parameter LlaMA 2 language model as the backbone~\cite{touvron2023llama2openfoundation}.
\subsubsection{Multimodal Input Encoding}

Given a video sequence of the speaker containing $T$ frames $\{\mathbf{I}_1, \mathbf{I}_2, \dots, \mathbf{I}_T\}$, we first extract comprehensive visual features. These features are fed into a projector to map them into the input space of the LLM, yielding a sequence of image tokens. 

To capture both fine-grained facial dynamics and global semantic context, we employ a dual-stream visual encoder. For each frame $\mathbf{I}_t$, we extract visual features using pre-trained DINO~\cite{oquab2023dinov2} and SigLIP~\cite{zhai2023sigmoid} models, given by 
\begin{equation}
    \mathbf{v}_t^{\text{dino}} = \text{DINO}(\mathbf{I}_t), \quad \mathbf{v}_t^{\text{siglip}} = \text{SigLIP}(\mathbf{I}_t),
\end{equation}
where $\mathbf{v}_t^{\text{dino}} \in \mathbb{R}^{d_v}$ captures pose and subtle expression details, while $\mathbf{v}_t^{\text{siglip}} \in \mathbb{R}^{d_v}$ encodes global affective semantics and social cues. The motivation for adopting this dual-branch encoder is that effective listener response necessitates both the capture of fine-grained facial details and head poses, as well as a high-level interpretation of emotional context.

The visual features are then projected and aggregated into image tokens $\mathbf{F}^{\text{vision}}$ through an multi-layer perceptron projector $\text{MLP}^{\text{vis}}$ to map them into the input space of the LLM, given by
\begin{equation}
    \mathbf{F}^{\text{vision}} = \{\text{MLP}^{\text{vis}}([\mathbf{v}_t^{\text{dino}}; \mathbf{v}_t^{\text{siglip}}]) \}_{t=1}^T,
\end{equation}
where $[\cdot;\cdot]$ denotes concatenation. 

The text content $\{\mathbf{L}_t\}_{1}^{T}$ is tokenized into text tokens $\mathbf{F}^{\text{text}}$ through the LLaMA~\cite{touvron2023llama2openfoundation} tokenizer,
\begin{equation}
    \mathbf{F}^{\text{text}} = \{\text{LLaMATokenizer}(\mathbf{L}_t)\}_{t=1}^T.
\end{equation}
Subsequently, these text tokens are concatenated with image tokens to serve as the unified multimodal input.

\subsubsection{Action De-Tokenizer}
To align continuous facial motions with the discrete output space of the  LLM, we implement a quantization strategy that maps continuous parameters into discrete tokens. Inspired by the work Rt-2~\cite{brohan2023rt2visionlanguageactionmodelstransfer}, we discretize each dimension of the facial actions into 256 bins. To determine the optimal quantization boundaries, we first sort the action values from the training corpus, and truncate the top and bottom 1\% to eliminate statistical outliers. The bins are then uniformly distributed over the remaining effective value range. This strategy effectively filters out noise and ensures that the limited representational capacity (i.e., the 256 bins) is concentrated within the valid motion interval, thereby significantly enhancing the modeling granularity for subtle micro-expressions and head poses. Finally, the discrete action tokens $\mathbf{F}^{\text{action}}$ predicted by the LLM are computed by
\begin{equation}
    \mathbf{F}^{\text{action}} = \text{LLM}([\mathbf{F}^{\text{vision}};\mathbf{F}^{\text{text}}]).
\end{equation}
$\mathbf{F}^{\text{action}}$ is mapped back to the continuous action $\mathbf{A}_{1:T}$ to synthesize the listener's response via 
\begin{equation}
    \mathbf{A}_{1:T} = \text{ActionDe-Tokenizer}(\mathbf{F}^{\text{action}}).
\end{equation}

\subsubsection{Supervised Fine-Tuning}

The first stage employs supervised fine-tuning to establish a reliable VLA mapping from multimodal inputs to facial actions. Given a dataset $\mathcal{D}_{\text{SFT}} = \{(\mathbf{I}_{1:T}, \mathbf{L}_{1:T}, \mathbf{A}_{1:T}^{*})^i\}_{i=1}^N$ containing aligned sequences of visual inputs $\mathbf{I}_{1:T}$, text inputs $\mathbf{L}_{1:T}$, and ground-truth facial actions $\mathbf{A}_{1:T}^{*}=\{\mathbf{a}_{1:T}^{\text{exp*}},\mathbf{a}_{1:T}^{\text{pose*}}\}$, we use cross-entropy loss $\mathcal{L}_{\text{cross}}$ to optimize the VLA parameters  $\theta$
\begin{equation}
    \begin{split}
            \mathcal{L}_{\text{pre}} =    \frac{\lambda_{\text{exp}}}{T} \sum_{t=1}^{T}  \mathcal{L}_{\text{cross}}(\mathbf{a}_{t}^{\text{exp*}} , \mathbf{a}_{t}^{\text{exp}})   + \\   \frac{\lambda_{\text{pose}} }{T} \sum_{t=1}^{T}  \mathcal{L}_{\text{cross}}(\mathbf{a}_{t}^{\text{pose*}} , \mathbf{a}_{t}^{\text{pose}})),
    \end{split}
\end{equation}
where $\lambda_{\text{exp}}$ and $\lambda_{\text{pose}}$ balance the contribution of expression and pose terms. To ensure temporal coherence and identity consistency, we incorporate additional regularization terms as
{\small
\begin{equation}
    \mathcal{L}_{\text{temp}} =\frac{1}{N}\sum_{i=1}^{N} \frac{1}{T-1} \sum_{t=1}^{T-1} (\|\mathbf{a}_{t+1}^{\text{exp}} - \mathbf{a}_t^{\text{exp}}\|_2^2+\|\mathbf{a}_{t+1}^{\text{pose}} - \mathbf{a}_t^{\text{pose}}\|_2^2).
\end{equation}
}
We yield the overall loss function
\begin{equation}
    \theta_{\text{SFT}}^* = \arg\min_{\theta} \big[ \mathcal{L}_{\text{pre}} + \lambda_{\text{temp}} \mathcal{L}_{\text{temp}}  \big], \label{equ:sft_loss}
\end{equation} 
where $\lambda_{\text{temp}}$ is the trade-off parameter. The model trained through Eq. (\ref{equ:sft_loss}) serves as a strong initial policy $\pi_{\theta_{\text{SFT}}^*}$ that produces visually coherent and identity-consistent facial responses, providing a solid foundation for subsequent human-feedback optimization.

\subsection{Human-Feedback Reinforcement Learning}

\noindent\textbf{Human Feedback Data Collection:} To construct the preference dataset $\mathcal{D}_{\text{pref}}$, we first utilize the initial policy $\pi_{\theta_{\text{SFT}}^*}$ to generate $N = 4$ candidate listener actions $\{\mathbf{A}_{1:T}^j\}_{j=1}^N$ for each speaker multimodel inputs $\mathbf{S}_{1:T}$. 
This process yields a set of candidate trajectories $\{(\mathbf{S}_{1:T}, \mathbf{A}_{1:T}^j)\}_{j=1}^N$. To provide a high-quality reference, we explicitly incorporate the ground-truth action sequence $\mathbf{A}^*_{1:T}$ into this set, forming a candidate group $ \{(\mathbf{S}_{1:T}, \mathbf{A}_{1:T}^1), \dots, (\mathbf{S}_{1:T}, \mathbf{A}_{1:T}^N), (\mathbf{S}_{1:T}, \mathbf{A}^*_{1:T})\}$. The collection of these groups constitutes our initial dataset. We also render the trajectories $(\mathbf{S}_{1:T}, \mathbf{A}_{1:T}^j)$ into the speaker-listener interaction videos $\tau^j$ to allow annotators to intuitively observe the complete speaker-listener dyadic interaction.

To ensure the reliability of human preference assessment, we establish a rigorous annotation criteria. Annotators are tasked with evaluating the listener's responsiveness across four rating functions: Empathy($\cdot$), Appropriateness($\cdot$), Engagement($\cdot$), and Naturalness($\cdot$). Based on these ratings, the final preference score is formulated as
{\small
\begin{equation}
    \begin{split}
        r(\tau^j) = & \alpha_{\text{emp}}\text{Empathy}(\tau^j) + \alpha_{\text{app}}\text{Appropriateness}(\tau^j) \\
        & + \alpha_{\text{eng}}\text{Engagement}(\tau^j) + \alpha_{\text{nat}}\text{Naturalness}(\tau^j).
    \end{split}
    \label{eq:10}
\end{equation}
}

The weighting coefficients $\alpha_{\text{emp}}, \alpha_{\text{app}}, \alpha_{\text{eng}}, \alpha_{\text{nat}} \in [0,1]$ balance the relative importance, with $\sum \alpha = 1$ ensuring normalized rating. Please refer to the \textit{Supplementary Material} for the detailed definitions and grading standards of each rating function.  We obtain a preference score for each trajectory within every candidate group $\mathcal{G} = \{(\mathbf{S}_{1:T}, \mathbf{A}_{1:T}^1,r^1), \dots, (\mathbf{S}_{1:T}, \mathbf{A}_{1:T}^N,r^N), (\mathbf{S}_{1:T}, \mathbf{A}^*_{1:T},r^*)\}$ by using  Eq.~(\ref{eq:10}).
\subsubsection{Reward Function}

Upon obtaining the scores for each trajectory, we rank the candidates within each group. Specifically, we designate the highest-ranked trajectory as the preferred response ($\mathbf{A}_{1:T}^w$) and the lowest-ranked trajectory as the dispreferred response ($\mathbf{A}_{1:T}^l$). This selection criteria ensures that $\mathbf{A}_{1:T}^w$ demonstrates superior contextual comprehension, timely responsiveness, and behavioral authenticity compared to $\mathbf{A}_{1:T}^l$, providing a high-confidence signal for alignment. 

Subsequently, we construct the final preference dataset $\mathcal{D}_{\text{pref}} = \{(\mathbf{S}_{1:T}, \mathbf{A}_{1:T}^w, \mathbf{A}_{1:T}^l)^i\}_{i=1}^K$ for Direct Preference Optimization (DPO)~\cite{rafailov2024directpreferenceoptimizationlanguage} training, where $\mathbf{A}_{1:T}^w$ and $\mathbf{A}_{1:T}^l$ denote the preferred(Pre) and dispreferred(Dispre) actions, respectively.

\subsubsection{Reinforcement Learning}

We use the DPO algorithm to drive the reinforcement learning.
We initialize the policy $\pi_\theta$ with the supervised fine-tuned model $\pi_{\text{ref}} = \pi_{\theta_{\text{SFT}}^*}$ and freeze a copy of $\pi_{\text{ref}}$ to serve as the reference distribution. 
The loss function is defined as
{\small
\begin{equation}
    \begin{split}
        \mathcal{L}_{\text{policy}} =  - \mathbb{E} \Bigg[ \log \sigma \bigg( & \beta \log \frac{\pi_\theta(\mathbf{A}_{1:T}^w \mid \mathbf{S}_{1:T})}{\pi_{\text{ref}}(\mathbf{A}_{1:T}^w \mid \mathbf{S}_{1:T})} \\ 
        & - \beta \log \frac{\pi_\theta(\mathbf{A}_{1:T}^l \mid \mathbf{S}_{1:T})}{\pi_{\text{ref}}(\mathbf{A}_{1:T}^l \mid \mathbf{S}_{1:T})} \bigg) \Bigg],
    \end{split}
    \label{equ:grpo} 
\end{equation}
}

where $\sigma$ is the sigmoid function, and $\beta$ is a hyperparameter controlling the strength of the KL-divergence constraint.

\section{Experiments}
\begin{table*}[htbp]
\small
\setlength{\tabcolsep}{10pt} 
\caption{Quantitative comparison on the L2L-trevor dataset. }
\centering
\renewcommand{\arraystretch}{1.3}
\fontsize{9pt}{10pt}\selectfont
\begin{tabular}{l|ccccccc}
\hline
Method & L2 $\downarrow$ & FD $\downarrow$ & Variation & Diversity & P-FD $\downarrow$ & L2 Affect($10^2$) $\downarrow$  \\
\hline
GT & &  & 0.1254 & 2.5427 &  \\
\hline
Random & 0.6852 & 33.5481 & 0.1458 & 2.7813 & 34.5579  & 12.4864 & \\
NN & 0.6047 & 28.4186 & 0.0914 & 2.1486 & 26.7514 & 10.6842 &  \\
LM-listener~\cite{NgSKKDG23} & 0.4345 & 17.6299 & 0.1189 & 2.9374 & 19.1583 & 6.3992 & \\
MMLHG~\cite{lai2025llm} & \textbf{0.2910} & \underline{10.0949} & 0.0704 & 2.2960 & 11.3908 & 2.6575 &  \\
\hline
    \textbf{Ours (SFT)} & \underline{0.3015} & \textbf{9.1473} & 0.0814 & 2.1756 & \underline{11.2975} & \underline{2.5724} &  \\
\textbf{Ours (SFT+RL)} & 0.3129 & 10.2385 & 0.0919 & 2.3754 & \textbf{10.8247} & \textbf{2.4842} & \\
\hline

\end{tabular}

\label{tab:l2l_results}
\end{table*}

\begin{table*}[htbp]
\small
\renewcommand{\arraystretch}{1.3}
\setlength{\tabcolsep}{10pt} 
\caption{Quantitative comparison on the Realtalk dataset.}
\centering
\fontsize{9pt}{10pt}\selectfont
\begin{tabular}{l|ccccccc}
\hline
Method & L2 $\downarrow$ & FD $\downarrow$ & Variation & Diversity & P-FD $\downarrow$ & L2 Affect($10^2$) $\downarrow$   \\
\hline
GT & &  & 0.0478 & 1.7250 &  \\
\hline
Random & 0.3749 & 17.2541 & 0.0549 & 1.9452 & 18.4436  & 24.1739  \\
NN & 0.3147 & 16.0572 & 0.0496 & 1.6241 & 15.9657 &  20.1478 \\
LM-listener~\cite{NgSKKDG23} & 0.2416 & 10.8423 & 0.0402 & 1.6148 & 10.5483 & 12.2730\\
MMLHG~\cite{lai2025llm} & 0.1021 & 3.7914 & 0.0234 & 1.2643 & 3.8145 & 6.0427 \\
\hline
\textbf{Ours (SFT)} & \textbf{0.0824} & \textbf{3.2425} & 0.0371 & 0.9142 & \underline{3.8036} & \underline{4.5207}\\
\textbf{Ours (SFT+RL)} & \underline{0.0973} & \underline{3.5842} & 0.0421 & 1.2457 & \textbf{3.7914} & \textbf{4.3531} & \\
\hline

\end{tabular}
\label{tab:realtalk_results}
\end{table*}

\subsection{Experimental Setup}

\subsubsection{Datasets}

We evaluate our method on two interaction datasets: \textit{L2L-trevor}~\cite{ng2022learning}, and \textit{RealTalk}~\cite{geng2023affectivefacesgoaldrivendyadic}. L2L-trevor is a single-listener dataset originally introduced in the work~\cite{ng2022learning}. Realtalk dataset is a more extensive corpus involving diverse speaker-listener pairs, and comprises 692 dialogue videos, totaling 115 hours of conversational interaction. These datasets collectively enable thorough evaluation of facial expression generation across different interaction scenarios. For detailed procedures regarding the datasets processing, please refer to the \textit{supplementary material}.

\subsubsection{Baseline Methods}
Following the evaluation protocols established in ~\cite{lai2025llm}, we compare our approach with four distinct baselines: Random, Nearest Neighbor (NN), LM-listener~\cite{NgSKKDG23}, and MMLHG~\cite{lai2025llm}, where the MMLHG is the current state-of-the-art multimodal framework that synthesizes listener reactions by fusing conversational content with the speaker’s facial FLAME parameters.

\subsubsection{Evaluation Metrics}

We employ comprehensive metrics to assess different aspects of generated facial expressions. The \textit{L2 distance} quantifies reconstruction accuracy: $\text{L2} = \frac{1}{T}\sum_{t=1}^T \|\mathbf{A}_t - \mathbf{A}_t^*\|_2^2$. The \textit{Fréchet Distance (FD)} measures distribution similarity between generated and real facial motions.
The \textit{Paired FD (P-FD)} evaluates motion quality while preserving temporal alignment: $\text{P-FD} = \frac{1}{N}\sum_{i=1}^N \text{FD}(\mathbf{A}_i^*, \mathbf{A}_i)$. The \textit{L2 Affect for synchrony (L2 Affect)} quantifies the accuracy of the predicted listener emotional affect across the sequence: $\text{L2 Affect} = \frac{1}{N} \sum_{i=1}^{N} (\bar{v}_{i} - \bar{v}_{i}^{GT})^{2}$, where $v$ denotes the facial emotional affect for the i-th frame, as estimated by the pre-trained EMOCA~\cite{danecek2022emocaemotiondrivenmonocular} model.The Variation and Diversity quantify the richness and dynamic properties of the generated facial actions.

\subsection{Quantitative Evaluation}

\subsubsection{Comparison with State-of-the-Art Methods}

Table~\ref{tab:l2l_results} and Table~\ref{tab:realtalk_results} present the comprehensive quantitative results across the two datasets, respectively. The results in Table~\ref{tab:l2l_results}  are taken from the original paper, while the results in Table~\ref{tab:realtalk_results} are our re-implemented version. We conducted this re-implementation to ensure a fair comparison on the Realtalk~\cite{geng2023affectivefacesgoaldrivendyadic} dataset, as the LM-listener\cite{NgSKKDG23} method was not originally evaluated on this benchmark, and the official source code for MMLHG~\cite{lai2025llm} is currently unavailable. 

Overall, our method demonstrates superior performance compared to existing methods. Specifically, our SFT model (i.e., our VLA model trained via supervised fine-tuning) excels in motion fidelity, achieving the lowest FD scores of 9.1473 (vs. MMLHG's 10.0949) on L2L-trevor and 3.2425 (vs. MMLHG's 3.7914) on Realtalk. Furthermore,  the SFT model attains an L2 error of 0.0824 on Realtalk, significantly outperforming the  values of 0.1021 and 0.2416 obtained by MMLHG and LM-listener  methods, respectively, which confirms its precision in capturing authentic facial dynamics.

Most importantly, the integration of the reinforcement learning stage (i.e., SFT+RL) drives substantial improvements in semantic and emotional alignment. As shown in Table 2, our SFT+RL model achieves the best L2 Affect score of 4.3531, surpassing both the SFT baseline (4.5207) and MMLHG (6.0427) by a wide margin. Similarly, on the L2L-trevor dataset shown in Table 1, our SFT+RL model records the lowest P-FD of 10.8247, improving upon the SFT model's 11.2975.  While the RL model exhibits a marginal increase in FD (e.g., 3.5842 on Realtalk and 10.2385 on L2L-trevor) indicating larger reconstruction errors, its superior performance in affective metrics validates that our reinforcement learning strategy effectively prioritizes social appropriateness and emotional consistency over simple geometric reconstruction.

\begin{figure*}[htbp]
\centering
\includegraphics[height=0.25\textheight]{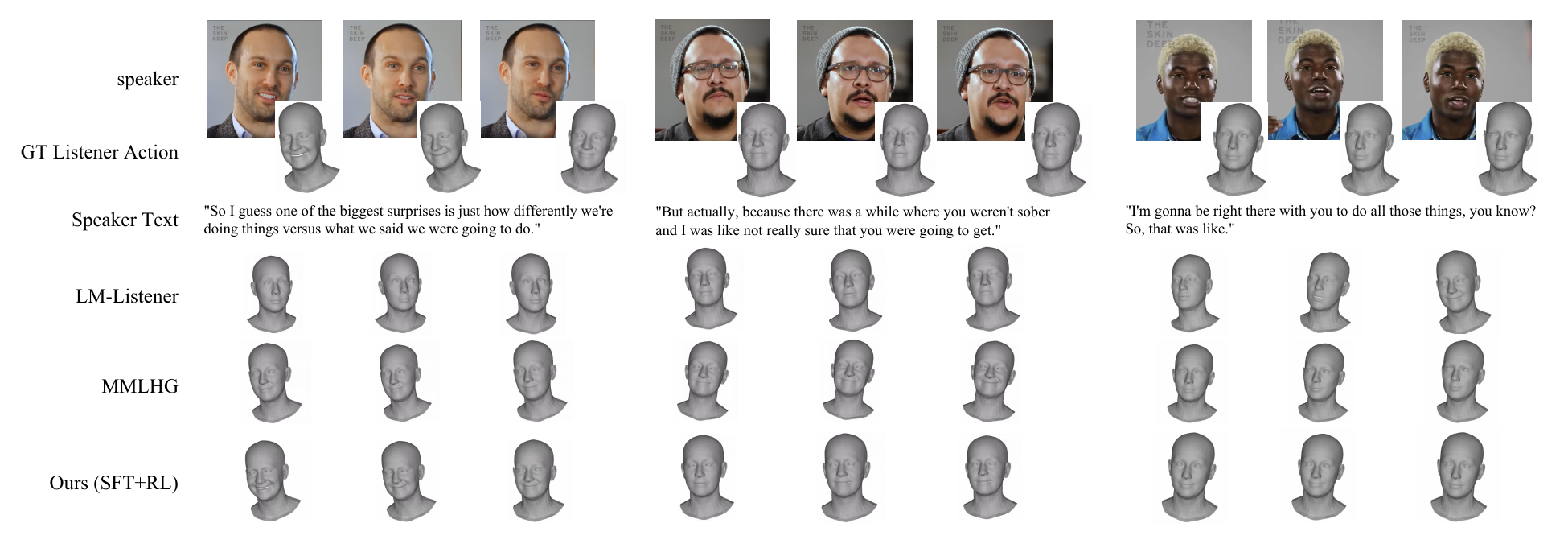}
\caption{Qualitative comparison of generated listener expressions on the Realtalk dataset. Our method produces more appropriate and emotionally aligned responses compared to baseline methods.}
\label{fig:qualitative}
\end{figure*}

\subsection{Qualitative Evaluation}


We visualize the listener responses generated by our method and baselines alongside the ground truth in Figure~\ref{fig:qualitative}. As observed in the left column, when the speaker expresses joy while discussing ``biggest surprises'', the LM-Listener~\cite{NgSKKDG23} fails to react, producing a neutral and apathetic face. MMLHG~\cite{lai2025llm} captures some positive sentiment, while our method also generates a more natural and intense smile, demonstrating superior emotional synchrony with the speaker's state.

The middle column in Figure~\ref{fig:qualitative} highlights the advantage of our preference alignment strategy. The speaker discusses a serious topic (``weren't sober''), necessitating a somber or attentive listener reaction. MMLHG exhibits a ''hallucinated'' positive emotion, generating an inappropriate smile that violates social norms. In contrast, our model correctly interprets the context and synthesizes a serious, attentive expression. As seen in the right column, while our model aligns well with the speaker's emotion, LM-listener fails to produce emotionally consistent expressions. These results confirm that our method not only mimics facial motions but also understands social appropriateness, avoiding the ``generic positivity'' bias often found in SFT-based models. 

\subsection{User Study}

\begin{table}[t]
\centering
\renewcommand{\arraystretch}{1.3}
\setlength{\tabcolsep}{5pt}
\caption{User study results (Mean Opinion Score, 1-5 scale). Our method achieves highest ratings across all aspects.}
\begin{tabular}{l|cccc}
\hline
Method & App. & Emp. & Eng. & Nat.  \\
\hline
LM-listener~\cite{NgSKKDG23} & 2.7 & 3.1 & 3.4 & 2.9 \\
MMLHG~\cite{lai2025llm} & 3.0 & 3.3 & 3.5 & 3.1 \\
\hline
\textbf{Ours (SFT)} & \underline{3.2} & \underline{3.4} & \underline{3.7} & \underline{3.3}  \\
\textbf{Ours (SFT+RL)} & \textbf{4.5} & \textbf{4.1} & \textbf{4.2} & \textbf{4.5}  \\
\hline
\end{tabular}

\label{tab:user_study}
\end{table}

To evaluate the perceptual quality and social appropriateness of the generated listener responses, we conduct a comprehensive user study involving 25 participants (aged 19-31, all university students or graduates). Each participant is presented with a series of videos. Each contains a speaker's video along with four corresponding listener responses generated by  the two baseline models, our SFT model, and our SFT+RL model. Participants are asked to rate the responses across four dimensions of Appropriateness (App.), Empathy (Emp.), Engagement (Eng.), and Naturalness (Nat.).

The quantitative results are summarized in Table \ref{tab:user_study}. Our SFT+RL method achieves significantly higher scores across all categories. Notably, the RL stage yields substantial gains over the SFT baseline, boosting Appropriateness from 3.2 to 4.5 and Empathy from 3.4 to 4.1. Furthermore, our method outperforms the strongest baseline MMLHG by a wide margin (e.g., +1.5 in Appropriateness and +0.8 in Empathy). These results confirm that our reinforcement learning strategy, guided by human feedback, effectively aligns the generated facial expressions with human social preferences, ensuring they are not only natural but also socially adept and empathetic. The implementation details of User study are provided in the \textit{supplementary materials}. 

\subsection{Ablation Study}

\begin{table}[t]
\centering
\renewcommand{\arraystretch}{1.5}
\setlength{\tabcolsep}{2.5pt}
\caption{Ablation study on the Realtalk dataset. }
\begin{tabular}{l|cccc}
\hline
\textbf{Method} & L2 $\downarrow$ & FD $\downarrow$ & P-ID $\downarrow$ & L2 Affect($10^2$) $\downarrow$ \\
\hline
Full (Ours) & 0.0973 & 3.5842 & \textbf{3.7914} & \textbf{4.3531} \\
\hline
Random-Prefer & 0.3142 & 12.3549 & 12.0354 & 15.2463 \\
SFT-Preferred  & 0.1132 & 3.6791 & 3.9165 & 5.519 \\
SFT-Only  & \textbf{0.0824} & \textbf{3.2425} & 3.8036 & 4.5207 \\
\hline
\end{tabular}

\label{tab:ablation}
\end{table}

We validate our design choices by comparing the full model against three variants: (1) SFT-Only, where the stage-one model is used   directly; (2) Random-Prefer, where DPO is used with random preference labels; and (3) SFT-Preferred, which uses standard SFT on the labeled trajectories with human preference.

As shown in Table~\ref{tab:ablation}, comparing Full (Ours) with SFT-Only reveals a trade-off: while SFT-Only yields lower reconstruction errors (L2/FD) by strictly imitating ground truth, our Full model achieves superior P-ID and L2 Affect scores. This significant confirms that reinforcement learning shifts the optimization focus from geometric reconstruction to social and emotional alignment. Furthermore, the degradation in Random-Prefer validates the necessity of accurate human feedback. Besides, our method outperforms SFT-Preferred, demonstrating that the contrastive DPO objective, which learns to distinguish between preferred and dispreferred behaviors, demonstrating more effective than straightforward supervision on positive samples alone.

\section{Conclusion}
In this paper, we present a novel method of facial expression generation aligned with human preference for dyadic interactions. Our method can effectively incorporate human feedback to ensure that generated listener expressions are contextually and emotionally congruent with the speaker's cues. Our method frames expression generation as an action learning process within an identity-independent parameter space, which can make the collection of unbiased human feedback focused on expressive quality. We introduce a vision-language-action model trained via supervised fine-tuning and refined by a human-feedback reinforcement learning strategy, which can dynamically produce appropriate facial responses. Extensive experiments on two dyadic benchmarks can demonstrate that our method outperforms state-of-the-art methods.

\bibliographystyle{IEEEtran}
\bibliography{ijcai26}

\end{document}